# Investigating Consistency in Query-Based Meeting Summarization:
# A Comparative Study of Different Embedding Methods

Chen, Jia-Chen (Oscar)
*School of Information Management*
*National Central University (NCU)*
Taoyuan, Republic of China R.O.C
(Taiwan)
h194420kk2@gmail.com

Guillem Senabre
*School of Eletronic Engineering*
*Unversitat Polièctnica de Catalunya (UPC)*
Barcelona, Spain
guillemsenabre@gmail.com

Allane Caron
*School of Computer Science*
*Université de technologie de Belfort-Montbéliard (UTBM)*
Sevenans, France
allanecaron21@gmail.com

*Abstract* — *With more and more advanced data analysis techniques emerging, people will expect these techniques to be applied in more complex tasks and solve problems in our daily lives. Text Summarization is one of famous applications in Natural Language Processing (NLP) field. It aims to automatically generate summary with important information based on a given context, which is important when you have to deal with piles of documents. Summarization techniques can help capture key points in a short time and bring convenience in works. One of applicable situation is meeting summarization, especially for important meetings that tend to be long, complicated, multi-topic and multi-person. Therefore, when people want to review specific content from a meeting, it will be hard and time-consuming to find the related spans in the meeting transcript. However, most of previous works focus on doing summarization for newsletters, scientific articles…etc, which have a clear document structure and an official format. For the documents with complex structures like transcripts, we think those works are not quite suitable for meeting summarization. Besides, the consistency of summary is another issue common to be discussed in NLP field. To conquer challenges of meeting summarization, we are inspired by* "QMSum: A New Benchmark for Query-based Multi-domain Meeting Summarization" *proposed by Microsoft and we also propose our Locater model designed to extract relevant spans based on given transcript and query, which are then summarized by Summarizer model. Furthermore, we perform a comparative study by applying different word embedding techniques to improve summary consistency. The dataset we used can be downloaded from Github:* https://github.com/Yale-LILY/QMSum, *and our code is available at* https://github.com/BakiRhina/Locator.

## I. Introduction

Text Summarization is one of the famous challenges in Natural Language Processing (NLP). There are various previous works have provided contributions to this task. However, as NLP techniques advance, an increasing number of potential applicable situations are being proposed. Query-based meeting summarization could be one of interesting challenge in this field. It aims to help people capture the points of meeting by asking query. Therefore, the model should be able to summarize the information related to the query based on the given meeting transcript.

Our project is inspired by an article authored by Microsoft, "QMSum: A New Benchmark for Query-based Multi-domain Meeting Summarization".[1] This paper introduces a comprehensive dataset designed to train models to perform query-based multi-domain meeting summarization. The authors suggest several potential avenues for future work, particularly focusing on improving the consistency between the generated summaries and the corresponding queries, as determined by the reference summaries. Therefore, our primary objective in this project is to generate a summary from given transcript and enhance the consistency of summary.

Given that Microsoft provided little code for us to leverage, we found ourselves recreating most of it. As such, certain aspects of our project yielded improvements in performance, while others, typically those that were more complex, resulted in limited advancements.

In this project, we build upon the foundation laid by Microsoft's work. Using the same dataset, we strive to develop a similar model, but with improved performance in enhancing summary consistency. Their model's architecture comprises two parts: a locator, which identifies the sections of the meeting that correlate with the associated question, and a summarizer, which summarizes the extracted sections of the meeting.

We adopted the similar architecture, but with different methods, models, and parameters, since we were not provided with any pre-existing code.

## II. Related work

### A. Text Summarization

Text Summarization is a common NLP task, which is to extract the important information from the given content and make a summary. Most of prior work focus on document summary [2], [3], [4], [5], [6] and try to do summarization with document datasets. For example, CNN/DailyMail [7], Newsroom [8] are News dataset, also, there are some previous works concentrate on other types of summarization, [9] are dedicated to generating a discourse-aware summary for scientific articles, [10] are dedicated to generating a query-focused multi-document summary (qMDS), which targets the use-case of generating actionable instructions from a set of sources, [1] are dedicated to generating a summary based on a meeting transcript. Our experiment mainly focuses on meeting summarization, a more challenging task compared to news summarization.

### B. Query-based Summarization

Query-based summarization aims to generate a brief summary according to a source document and a given query, e.g [11] derived the encode-attend-decode paradigm to highlight those points that are relevant in the context for a given query. Meeting is also a genre of discourses where query-based summarization could be applied, [12] propose a Ranker-Generator framework to perform Query-Focused Meeting Summarization (QFMS), [1] develop a **Locate-** **then-Summarize** framework to perform query-based meeting summarization. By employing these query-based approaches, the generated summary can contain more precise and relevant information related to the specific queries. This enables the summarization model to produce summaries that cater to the reader's needs in a more adaptable manner.

*C. Meeting Summarization*

Meeting Summarization is a novel branch of text summarization task, which takes meeting transcripts as the content and summarizes the meeting. It is a useful and interesting application which can help people to catch the points from the meetings efficiently. Meeting summarization is really different with the document summarization due to the structure of original content. For the document, people are used to writing in a specific structure and mentioning important points (e.g subjects, intention, topics…etc) in the beginning of the document. However, there is no specific structure for a meeting transcript so summarizing a meeting would be more challenging than summarizing a document.

Even so, there are still some previous work take it as a challenge, [13] propose an unsupervised graph-based framework for abstractive meeting speech summarization, [14] design a hierarchical structure to accommodate long meeting transcripts and a role vector to depict the difference among speakers. Furthermore, there are some research focus on integrating the query-matching mechanism with meeting summarization [1], [11], [12], [15], which can generate a summary only relevant to the given queries instead of the whole meeting and help readers to catch the specific points in the meeting.

*D. Summarization Consistency*

According to [16], there are 4 dimensions to evaluate if a summary is good or bad. Consistency is one of the aspects that should be evaluated. Most previous studies define consistency as factual alignment between the summary and the source. If there is a summary with low consistency, the neural hallucination will happen, which means a model produces fluent summaries and frequently contains false or unsupported information. However, the definition of consistency in previous works are obviously not comprehensive enough for a query-based summarization. Here, we define consistency as query-related factual alignment between the summary and the source.

## III. DATASET

The dataset we are using to train our Locator, is the same as the one provided by Microsoft, "QMSum". This dataset can be found in the due github repository https://github.com/Yale-LILY/QMSum, and consists of 1808 query-summary pairs over 232 meetings in multiple domains. Of these 232 meetings, 162 are dedicated to the training set that will be used to train our Locator. The other meetings are reserved to the test and validation dataset, that will be used to evaluate the performance of our model, Locator and Summarizer.

To fully comprehend our experiment, we will delve into the details of how this dataset was constructed in this section. In the realm of query-based meeting summarization, very few benchmark datasets exist, even fewer are multi-domain. This unique aspect makes this dataset a particularly intriguing new benchmark for such applications.

After transcribing numerous meetings, annotators manually created general questions for each meeting, followed by their associated summaries. They repeated this process for more specific questions as well. Thus, for every meeting, we have both general and specific questions along with their associated summaries, which facilitated the training of several models subsequently.

The following diagram (Fig. 1), extracted directly from the cited article, presents the process of creating this dataset.

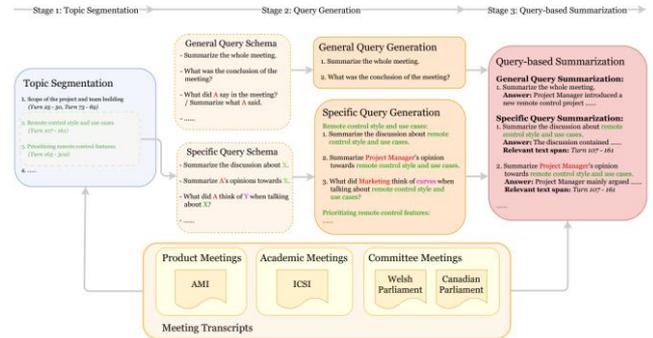

Fig. 1. The creation process of describe QMSum dataset.

To get a little bit more familiar with the dataset we have in front of us, a sample's being shown below, where we can see the meeting. (Fig. 2)

```
{
    "topic_list": [
        {
            "topic": "Introduction of petitions and prioritization of governmental matters",
            "relevant_text_span": [["0","19"]]
        },
        [...]
    ],
    "general_query_list": [
        {
            "query": "Summarize the whole meeting.",
            "answer": "The meeting of the standing committee took place to discuss matters pertinent to the Coronavirus pandemic. The main issue at stake was to ..."
        },
        [...]
    ],
    "specific_query_list": [
        {
            "query": "Summarize the discussion about introduction of petitions and prioritization of government matters.",
            "answer": "The Chair brought the meeting to order, announcing that the purpose of the meeting was to discuss COVID-19 's impact on Canada. Five petitions were presented ...",
            "relevant_text_span": [["0","19"]]
        },
        [...]
    ],
    "meeting_transcripts": [
        {
            "speaker": "The Chair (Hon. Anthony Rota (NipissingTimiskaming, Lib.))",
            "content": "I call the meeting to order.  Welcome to the third meeting of the House of Commons Special Committee on the COVID-19 Pandemic ..."
        },
        [...]
    ]
}
```

Fig. 2. The content of meeting transcript in QMSum.

As you can see, inside the dataset we find 4 main features, "topic_list", "general_query_list", "specific_query_list" and "meeting_transcripts". Specifically we are going to focus on the "specific_query_list" which provides a specific query, the answer of that query in the meeting and the relevant spans of the transcript where the answer can be found, and the "transcripts" where a "speaker" and "content" dictionaries can be found. The combination of a "speaker" and a "content", where the content is what the speaker is saying, is called a turn. So each meeting has a transcript containing n turns.

We preprocess these meetings with a specific and custom text preprocessing, that not only will provide the basic preprocessing, but also will remove the non-important words the voice recognition generates, such as 'vocalsound', 'disfmarker', 'nonvocalsound', for instance. This preprocessing can be found in the "PREPROCESSING" section of the Locator.ipynb.

Since the structure of the validation set and train set have slight but important differences, we created a custom preprocessing section for each one of them, that can easily be found on the Locator.ipynb.

## IV. METHODOLOGY

### A. Pre-trained Model Embedding

Here, we mainly discuss two types of pre-trained models based on transformers architecture [17]. According to lots of prior research, different embedding methods may cause different influences to the model. Therefore, there is some research focusing on improving the performance by employing different embedding to represent text, sentence, document…etc. [18] present a general framework to support cross-corpora language studies, which can efficiently find semantic differences across corporas. [19], [20], [21] also mention that some pretrained embedding models are widely used for NLP tasks because of benefits from large corpus and better representations for contextual information e.g BERT [22], RoBERTa [23], ELECTRA [24]...etc.

Furthermore, [25] indicates that the performance of summarization can be improved by using a better embedding for representation. Therefore, we will try to improve the consistency of summary by employing different embeddings. In the experiment, we will use BERT, RoBERTa, ELECTRA as our embedding methods, which have been proved to be useful in lots of NLP tasks.

### B. Model

#### 1) Locate-then-Summarize

To perform a query-based meeting summarization, various approaches have been proposed by other previous works. One of common approach is **Locate-then-Summarize** framework [1]. In this two-stage pipeline, the first step requires a model to locate the relevant text spans in the meeting according to the queries, and we call this model a Locator. The reason why we need a Locator here is, most existing abstractive models cannot process long texts such as meeting transcripts. So we need to extract shorter, query-related paragraphs as input to the following Summarizer which is another model for generating the final summary based on the given content from Locater.

#### 2) Locater

In the experiment, we build our Locater with a similar structure like [1] see Fig. 3. First, all the turns in the meeting transcript will be embedded utterance-by-utterance, then we take the concept of average word embedding (AWE) and use it in our case. Some research indicates that is an easy and universal way to represent a sentence, paragraph or even document [26]. Therefore, average utterance embedding will be used to represent the whole transcript document embedding in the experiment. Also, we take the same ways to embed the queries but not averages, because each query should be independent of each other.

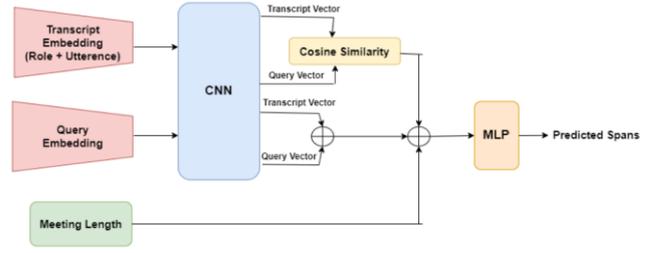

Fig. 3. The structure of our Locater model.

After the embedding process, both transcript embedding and query embedding will be put into a CNN model to extract local features and compress it into a shorter vector respectively, just like [27] use a CNN as an encoder to perform extractive summarization. Let $T_i$ represents transcript embedding for $i_{-th}$ meeting, $Q_{ij}$ represent query embedding for $j_{-th}$ query in $i_{-th}$ meeting, $S_{ij}$ represent span index for $j_{-th}$ query in $i_{-th}$ meeting.

$$T'_i = Max - pool(Conv1d(T_i)) \quad (1)$$

$$Q'_{ij} = Max - pool(Conv1d(Q_{ij})) \quad (2)$$

$$E^T_i = W_1\ T'_i + b_1 \quad (3)$$

$$E^Q_{ij} = W_1\ Q'_{ij} + b_1 \quad (4)$$

$Max - pool$ is max pooling computation. $Conv1d$ is 1d-convolution of CNN. $W_1$ and $b_1$ are learnable parameters.

Then, we concatenate the transcript and query vectors that CNN outputs, and take it as inputs of MLP to predict the index of <START, END> which means the start and end of the relevance spans. However, we find that the predicted results are quite unreasonable, some of <START, END> are even out of the range of meeting length. It is probably because Locator has never known the length of meeting and similarity between transcript and query during training, therefore, we decide to concatenate the meeting length and similarity value between transcript vector and query vector as two additional inputs features of MLP. Let $Similarity()$ represents similarity function, here, we apply cosine similarity as our similarity function. $L_i$ represent the length of meeting. As for MLP, we apply LeakyReLu as our activation function.

$$Sim_{ij} = Similarity(E^T_i, E^Q_{ij}) \quad (5)$$

$$E_{ij} = E^T_i \oplus E^Q_{ij} \oplus Sim_{ij} \oplus L_i \quad (6)$$

$$E'_{ij} = LeakyReLu(W_2\ E_{ij} + b_2) \quad (7)$$

$$S'_{ij} = Abs(W_3\ E'_{ij} + b_3) \quad (8)$$

$Sim_{ij}$ represent cosine similarity value between $E^T_i$ and $E^Q_{ij}$. Symbol $\oplus$ represent concatenation. $Abs()$ means absolute value. $W_2$, $W_3$, $b_2$, $b_3$ are learnable parameters. $S'_{ij}$ is the output of our Locater model, which is used to represent the predicted index of spans in <START, END> format.

With meeting length and cosine similarity, the predicted spans become much more reasonable. So the following experiment will train the Locator with meeting length and cosine similarity. After acquiring the predicted spans index $S'_{ij}$, we can further compute loss value with ground truth index.

$$Loss = f(S_{ij}, S'_{ij}) \qquad (9)$$

Let $f$ represent loss function, $S_{ij}$ and $S'_{ij}$ are true index and predicted index, respectively.

*3) Summarizer*

The second part of the model's architecture pertains to the summarizer. In the original paper, they utilized several models capable of summarizing texts, namely BART [28], PGNet [29], and HMNet [14]. For our experiment, we decided to utilize only BART and a version of BART fine-tuned on meeting summaries found on Hugging Face (which we will refer to as BART_MEETING). Here is the link to this model:
https://huggingface.co/knkarthick/MEETING_SUMMARY .

To accurately assess the performance of our summarizer and locator, we followed the same process as in the original paper (Fig. 4), i.e. first testing our summarizers on the gold spans (representing the meeting excerpts extracted by the annotators, which correspond to the question), and then testing our summarizers on the "located spans" (representing the meeting excerpts extracted by our locator(s)). This approach allows us to evaluate our summarizer and compare it with those mentioned in the original paper.

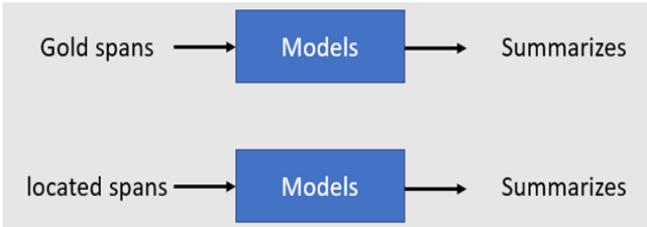

Fig. 4. The process of evaluation.

Moreover, for the two models used in our experiment to summarize spans, we decided to fine-tune them on the dataset to enhance their performance, leading us to 4 different summarizers.

Note: The models used have a maximum input capacity of 1024 tokens. As such, extracts exceeding this limit had to be truncated to be processed by the models. We maintained the same input format for the summarizers as in the original paper, that is, <s> Query </s> relevant spans </s>.

## V. EXPERIMENT

Our experiment is inspired by [1], and the objective is to improve the consistency of the final summary. According to lots of prior research, the embedding methods may cause significant influences to the performance of models. Therefore, we decided to conduct the experiment by employing different embedding methods (BERT [22], RoBERTa [23], ELECTRA [24]) to further investigate the impact of different embedding techniques and find out which one could help the model improve the summary consistency. The experiment process is shown in Fig. 5. The "Pretrained Model Embedding" module in Fig. 5 will apply different pretrained word embedding models. In the end, we will evaluate results of the Summarizer model.

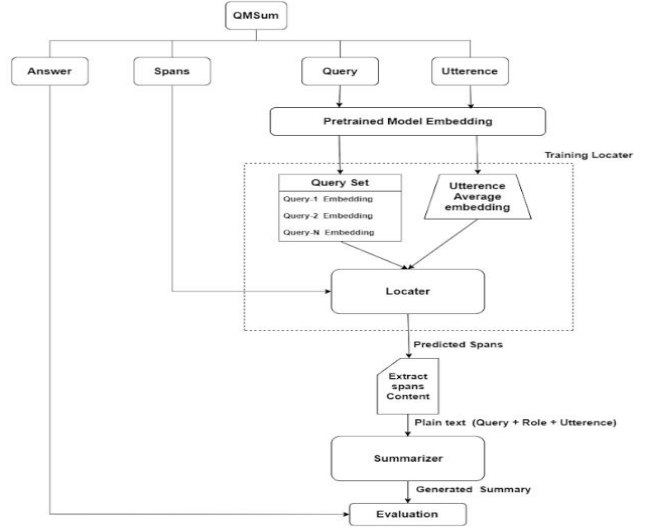

Fig. 5. The experiment process of our project.

## VI. EVALUATION

*A. Metrics*

We opted to use the same metric as in the Microsoft paper, specifically, the ROUGE [30] measure. These metrics provide a quantitative assessment of summary quality and help compare different summarization approaches. For our experiment, we will focus exclusively on R-1, R-2, and R-L measures, just like [1] used in their experiment.

**ROUGE (Recall-Oriented Understudy for Gisting Evaluation)** is a set of metrics used to evaluate the quality of automatic summaries and machine translation outputs. It includes three main variants:

(1) **ROUGE-N (R1, R2)**: Measures the overlap of n-grams (continuous sequences of n words) between the generated and reference summaries.

(2) **ROUGE-L**: Measures the longest common subsequence (LCS) between the generated and reference summaries, considering word order.

(3) **ROUGE-S**: Measures the overlap of skip-bigrams (pairs of non-adjacent words in the same order) between the generated and reference summaries.

*B. Result*

The following table (Fig. 6) presents the ROUGE scores for all the locators with summarizers. Also, we provide the performance from [1] at the top of the table. However, the prior performance cannot be strictly compared because it did not provide the original locator model, which means we have to build the locator model by ourselves and some differences in detail will be inevitable. According to the results obtained, the best score is for the MEETING_SUMMARIZE models and the gold spans, with an R-L of 33.1 for this model. This surpasses the best model from Microsoft (using gold spans), which scored 31.27. However, the scores with our locators are relatively modest.

An asterisk (*) indicates that the model has been fine-tuned on the QMSum dataset. On the right side of the table, we can observe the impact that fine-tuning has had on our models' performance. For example, using fine-tuning to generate summaries from the gold spans has, on average, improved the ROUGE scores by 80%.

As for ROBERTA, BERT, and ELECTRA, fine-tuning has respectively improved the scores by 50%, 71%, and 62%. Furthermore, we notice that the performance of the summaries on the relevant spans extracted by our three locators is far less consistent than on the gold spans (based on ROUGE scores). Indeed, even with the fine-tuned models, our summaries achieve an average ROUGE-L of 16.2, whereas with the gold spans, they achieve an average of 26.5, which is roughly 39% less.

Despite the overall low performance of the three locators, the one using ROBERTA yielded slightly superior scores.

| Spans (Locator) | Model (Summarizer) | R-1 | R-2 | R-L | R-1 | R-2 | R-L |
|---|---|---|---|---|---|---|---|
| Microsoft results | | | | | | | |
| Gold spans (without locator) | BART | 32.18 | 8.48 | 28.56 | | | |
| Relevant spans (their locator) | BART | 31.74 | 8.67 | 28.17 | | | |
| Our results | | | | | | | |
| Gold spans (without locator) | MEETING SUM | 24 | 5.8 | 21.4 | 56.7 | 127.6 | 54.7 |
| | BART | 22.8 | 5.4 | 20.2 | 58.8 | 125.9 | 54.5 |
| | | | | | Mean | | 79.7 |
| | MEETING SUM* | 37.6 | 13.2 | 33.1 | | | |
| | BART* | 36.2 | 12.2 | 31.2 | | | |
| | Mean | 30.2 | 9.2 | 26.5 | | | |
| Relevant spans (with our locator - Roberta) | MEETING SUM | 19.7 | 3.5 | 16.1 | 29.4 | 60.0 | 18.6 |
| | BART | 17.1 | 2.2 | 14 | 40.4 | 127.3 | 24.3 |
| | | | | | Mean | | 50.0 |
| | MEETING SUM* | 25.5 | 5.6 | 19.1 | | | |
| | BART* | 24 | 5 | 17.4 | | | |
| | Mean | 21.6 | 4.1 | 16.7 | | | |
| Relevant spans (with our locator - BERT) | MEETING SUM | 17 | 3.7 | 13.8 | 43.5 | 62.2 | 29.0 |
| | BART | 14.7 | 2 | 13.4 | 71.4 | 185.0 | 36.6 |
| | | | | | Mean | | 71.3 |
| | MEETING SUM* | 24.4 | 6 | 17.8 | | | |
| | BART* | 25.2 | 5.7 | 18.3 | | | |
| | Mean | 20.3 | 4.4 | 15.8 | | | |
| Relevant spans (with our locator - Electra) | MEETING SUM | 18.3 | 3.5 | 15 | 38.3 | 62.9 | 24.0 |
| | BART | 15.7 | 2 | 13.1 | 55.4 | 155.0 | 35.9 |
| | | | | | Mean | | 61.9 |
| | MEETING SUM* | 25.3 | 5.7 | 18.6 | | | |
| | BART* | 24.4 | 5.1 | 17.8 | | | |
| | Mean | 20.925 | 4.075 | 16.125 | | | |

* fine tuned with QMSum dataset

Fig. 6. The performance of our project.

## VII. CONCLUSION

There is no doubt that this dataset is comprehensive and highly effective for such a task. Our initial goal was to improve the consistency of our summaries, and based on the ROUGE scores, we have seen a slight enhancement in our summarizer's performance. Indeed, for all our summarizers (used with gold spans), we obtain an average ROUGE-L score of 26.46, with a clear difference between the non-fine-tuned models and those fine-tuned on the QMSum datasets. Specifically, the fine-tuned models show an average improvement (all ROUGE scores combined, all locators combined) of 65.7%.

As for the performance of our locators, we have an average ROUGE-L score for locators using ROBERTA, BERT, and ELECTRA, which are respectively 16.65, 15.83, and 16.1. Given that the original paper stated the average ROUGE-L when using a random locator is 11.76, this demonstrates the comparatively low performance of our locators.

Going forward, this study provides a strong foundation for future exploration. As we continue to improve upon the current models and explore new methodologies, we believe that we can further enhance the consistency of the summaries and the performance of the framework in following aspects.

(1) **Out-of-Range Spans** — During the prediction we find that the predicted indexes of relevant spans sometime will be out of range of meeting length, which means the locator model cannot locate the relevant spans for some queries. Therefore, if this out-of-range problem can be solved or improved, there are reasons to believe the performance of the whole framework will increase.

(2) **Consistency Metrics** — Although ROUGE scores are commonly used in summarization tasks. However, ROUGE only is obviously not enough for evaluating consistency. [31] provide a comprehensive overview of various metrics for summarization tasks which should be considered in the future.

Our results, despite some limitations, demonstrate promising potential and pave the way for continued advancements in the field of query-based multi-domain meeting summarization.

ACKNOWLEDGMENT

This project was completed in June 2023. It was amazing to see people from three different countries collaborate and complete this challenge together. During this period, we believe that what we gained from this project is not only technical knowledge but also valuable friendships. It was an amazing and interesting experience for all of us. Best regards to all the members of the project team.